\documentclass[sigconf]{acmart}
\usepackage{multirow}
\usepackage{graphicx}
\usepackage{booktabs}
\usepackage{color, colortbl}
\usepackage[font=small,skip=7pt]{caption}
\usepackage{enumitem}

\definecolor{Gray}{gray}{0.9}
\AtBeginDocument{%
  }

\acmConference[ICMR]{Conference}{June, 2024}{Phuket, Thailand}
  
\begin{document}


\title{CLIPping the Deception: Adapting Vision-Language Models for Universal~Deepfake~Detection}






\author{Sohail Ahmed Khan}
\affiliation{%
  \institution{University of Bergen}
  \city{Bergen}
  \country{Norway}}
\email{sohail.khan@uib.no}

\author{Duc-Tien Dang-Nguyen}
\affiliation{%
  \institution{University of Bergen}
  \city{Bergen}
  \country{Norway}}
\email{ductien.dangnguyen@uib.no}



\begin{abstract}
The recent advancements in Generative Adversarial Networks (GANs) and the emergence of Diffusion models have significantly streamlined the production of highly realistic and widely accessible synthetic content. As a result, there is a pressing need for effective general purpose detection mechanisms to mitigate the potential risks posed by deepfakes. In this paper, we explore the effectiveness of pre-trained vision-language models (VLMs) when paired with recent adaptation methods for universal deepfake detection. Following previous studies in this domain, we employ only a single dataset (ProGAN) in order to adapt CLIP for deepfake detection. However, in contrast to prior research, which rely solely on the visual part of CLIP while ignoring its textual component, our analysis reveals that retaining the text part is crucial. Consequently, the simple and lightweight Prompt Tuning based adaptation strategy that we employ outperforms the previous SOTA approach by \textbf{5.01\%} mAP and \textbf{6.61\%} accuracy while utilizing less than one third of the training data (200k images as compared to 720k). To assess the real-world applicability of our proposed models, we conduct a comprehensive evaluation across various scenarios. This involves rigorous testing on images sourced from \textbf{21} distinct datasets, including those generated by GANs-based, Diffusion-based and Commercial tools. \textcolor{red}{Code and pre-trained models will be made available:} \href{https://github.com/}{\color{red}{https://github.com/}}


\end{abstract}







\maketitle

\section{Introduction}
\label{sec:intro}

 
The internet is now flooded with synthetic images generated by deep neural networks, commonly known as "Deepfakes," thanks to technologies like Generative Adversarial Networks (GANs)~\cite{goodfellow2014generative, karras2020training} and Denoising Diffusion Probabilistic Models (DDPMs)~\cite{sohl2015deep, rombach2022high}. These powerful tools have become accessible to a wider audience due to open-source availability.

In response to this, researchers have been actively proposing novel methods for automatic detection of synthetic content~\cite{wang2020cnn, ciftci2020fakecatcher, chen2022self, ojha2023towards, wang2023dire}. However, a major issue with existing deepfake detection models is their limited ability to generalize across different data distributions~\cite{mirsky2021creation, chen2022self, ojha2023towards}.
Deepfake detection is typically posed as a supervised learning problem, where a deep neural network model is trained to differentiate between authentic (\textit{real}) and manipulated (\textit{fake}) images~\cite{zhu2021face, ojha2023towards, wang2020cnn}. However, a significant challenge arises: if the model is exclusively trained on a particular category of fake images, its performance may falter when confronted with novel types of manipulated images, i.e., the generalization dilemma~\cite{khan2023deepfake}.

\begin{figure}[t!]
  \centering
  \includegraphics[width=\linewidth]{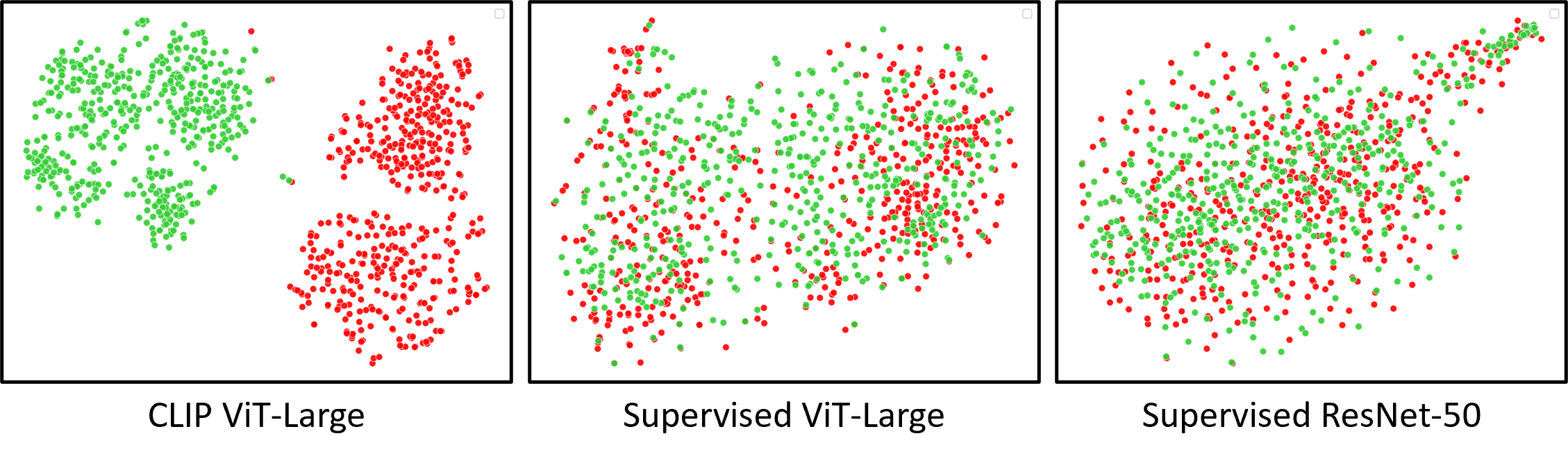}
\caption{Visualization of real (in red) and fake (in green) images utilizing t-SNE in the feature space of various image encoders. The feature space of CLIP demonstrates superior separation of real and fake image features as compared to other two supervised models.}
\label{fig:tsne}
\end{figure}

In~\cite{ojha2023towards}, \citeauthor{ojha2023towards} suggest that current detection models might be biased towards identifying certain types of fake images because they focus on easily detectable patterns found in those images. As a result, these models might miss out on the subtle features of real images, treating them as if they do not match the patterns learned from the fake images. In order to overcome this, the authors proposed to conduct classification using models that have trained on diverse range of images during their initial training, i.e., models that are not specifically trained for deepfake detection. They proposed to employ large vision-language models, in particular, CLIP (Contrastive Language-Image Pre-training)~\cite{radford2021learning} as a feature extraction model, and train a linear classification head on top for detecting deepfakes. They also observed that CLIP, even without undergoing specific training for classifying real and fake images, exhibits remarkable capability right from the start in discerning between authentic and fake images. Refer to Figure~\ref{fig:tsne} for details.


In~\cite{ojha2023towards}, \citeauthor{ojha2023towards} adapted CLIP for deepfake detection using linear probing, and the results they achieved showed strong generalization capabilities as compared to previous state-of-the-art~\cite{wang2020cnn} in detecting deepfakes. However, as highlighted in~\cite{zhou2022learning, kim2021adapt}, adapting CLIP through linear probing does not exploit its language component, and only relies on visual features, which can lead to sub-optimal performance. Our hypothesis is that by adapting CLIP using both the visual and text encoders, we can enhance detection performance, leading to a more effective and generalizable strategy for deepfake detection. In order to then verify our hypothesis we now raise this question: "Could combining CLIP's visual and textual capabilities further improve deepfake detection methods?" 

In pursuit of an answer, we delve into existing research literature focused on adapting Vision-Language Models (VLMs), specifically CLIP~\cite{radford2021learning}, for image classification tasks. For instance, Prompt Tuning~\cite{zhou2022learning} by \citeauthor{zhou2022learning} involves adapting a pre-trained CLIP model using language supervision. This method freezes the large CLIP model, and optimizes a small embedding treated as a prompt. In~\cite{gao2023clip}, CLIP Adapter is introduced, which adds a lightweight linear layer inside the CLIP model. During training, the large CLIP model remains frozen, while the smaller linear layer is optimized. Surprisingly, these promising strategies have not been explored in detecting deepfakes. 
The primary focus of our study is thus \textbf{to determine the most effective transfer learning strategy among various options for large vision-language models in the context of deepfake detection}. Moreover, we also pose questions such as, how various experimental conditions might impact the performance of the adopted strategies. This includes examining their ability to generalize to unseen data, performance when trained with limited real or fake image samples, robustness to different post-processing operations, and the impact of using a restricted amount of data for training.



To answer all these questions, we conduct an empirical analysis of the robustness of CLIP~\cite{radford2021learning} when trained using these strategies, and evaluate resulting models on data originating from varied distributions. Specifically, we take the pre-trained CLIP model, and train it for deepfake detection using four distinct strategies, including (1) Fine-tuning, (2) Linear Probing, (3) Prompt Tuning~\cite{zhou2022learning} and (4) training an Adapter Network~\cite{gao2023clip}. Following \cite{wang2020cnn} and \cite{ojha2023towards}, we employ ProGAN~\cite{karras2017progressive} as our training set. However, in contrast to these studies, we only use $200k$ images for training as compared to $720k$ images used by these two studies. We analyze our models on an extensive test set comprising of $21$ different GAN-based, Diffusion-based and Commercial image generators. Our approach achieves high classification performance while using less training data as compared to previous approaches.

Our contributions can be summarised as follows:
\begin{itemize}[leftmargin=*]
    \item We conduct an extensive empirical investigation into four distinct transfer learning strategies aimed at enhancing the adaptability and robustness of CLIP for deepfake detection, while taking inspiration from recent research on adapting large VLMs.
    \item Through experimentation, we illustrate that our chosen transfer learning strategies, notably Prompt Tuning, beats the current state-of-the-art~\cite{ojha2023towards} by a clear margin.
    \item We carry out few-shot experiments, illustrating excellent performance of our models even when exposed to only 32 $real/fake$ samples from each LSUN object category~\cite{yu2015lsun}, highlighting the effectiveness of the selected lightweight transfer learning strategies.
    \item Robustness analysis conducted in the presence of post-processing operations such as JPEG compression and Gaussian blurring.
    \item Analysis of the impact of training set size, demonstrating that CLIP-based detectors achieve solid performance even when trained using a smaller amount of data (20k real fake images).
    \item We plan on making the associated code and trained models open-source for the benefit of research community.
\end{itemize}

This paper is organized as follows. In Section~\ref{sec:relatedworks} we present a brief description of related works. In Section~\ref{sec:methodology} we introduce the problem background, our proposed deepfake detection workflows, and the datasets that we employ for evaluation of our models. In Section~\ref{sec:experiments} we elaborate in detail about the experiments we carried out for the sake of this study, and discuss the achieved results. Finally in Section~\ref{sec:conclusion} we conclude our study.

\section{Related Works}
\label{sec:relatedworks}
\subsection{Pre-trained Vision-Language Models}
Recent advancements in large-scale pre-trained models, which integrate vision and language capabilities, have showcased notable success across a variety of tasks encompassing both images and text~\cite{alayrac2022flamingo, radford2021learning, jia2021scaling}. The primary rationale driving the extensive adoption of these models lies in their interesting zero-shot capabilities and robustness to distribution shifts.

\citeauthor{radford2021learning} proposed Contrastive Language-Image Pre-training (CLIP), a large-scale model that exhibits robust zero-shot performance on several downstream tasks including image classification, optical character recognition, image text retrieval, and multiple other tasks~\cite{radford2021learning}. CLIP was pre-trained on a large scale dataset containing 400 million images, and their associated text captions. CLIP was pre-trained utilizing a contrastive loss, aiming to maximize the similarity between corresponding image and text captions compared to dissimilar pairs.

Moving away from the requirement of expensive data cleaning process similar to \citeauthor{radford2021learning}, \citeauthor{jia2021scaling}~\cite{jia2021scaling} utilized a large-scale noisy dataset containing one billion image-text pairs to pre-train their model. The model was comprised of dual-encoder architecture, which was tasked to align visual and language representations of image-text pairs through a contrastive loss. They showed that a large enough dataset can compensate for its noise, resulting in state-of-the-art representations even with such a straightforward learning approach.

\subsection{Transfer Learning}
Vision and language models like CLIP~\cite{radford2021learning} and ALIGN~\cite{jia2021scaling} offer interesting zero-shot capabilities on several different downstream tasks. Yet, to attain performance levels comparable to state-of-the-art models on these downstream tasks, these models require further fine-tuning on task-specific datasets. For example, even on a simple dataset like MNIST~\cite{lecun1998gradient}, the zero-shot CLIP model (ViT-B/16) which was tested in~\cite{kim2021adapt} achieved an accuracy of only 55\%.

However, it becomes apparent that fine-tuning full model on downstream dataset affects its robustness to distribution shifts~\cite{radford2021learning, wortsman2022robust}. In response to this challenge, several studies have introduced techniques to fine-tune large vision and language models. In~\cite{zhou2022learning} \citeauthor{zhou2022learning} proposed Context Optimization (CoOp), a fine-tuning strategy to adapt vision-language models similar to CLIP for downstream image classification tasks. CoOp injects learnable vectors to a textual prompt's context (either at the front, middle or end), which are optimized during fine-tuning by minimizing the classification loss, whereas, both the vision and text encoders of CLIP are kept frozen. \citeauthor{gao2023clip} introduced CLIP-Adapter~\cite{gao2023clip}, a bottleneck layer designed to learn new features during fine-tuning. Additionally, it employs a residual-style feature aggregation approach to seamlessly integrate the originally pre-trained CLIP features with the newly acquired ones, all while keeping CLIP model frozen itself. 



\subsection{Fake Image Generation and Detection}

Deep learning models for fake image generation have been with us for quite some time. \citeauthor{goodfellow2014generative} initially introduced Generative Adversarial Networks (GANs), a neural network architecture for unconditional fake image generation~\cite{goodfellow2014generative}. Seminal works were targeted on for example, improved training process of GANs~\cite{salimans2016improved, gulrajani2017improved, karras2020training}, improving quality and diversity of the generated images~\cite{karras2017progressive, karras2020analyzing} and conditional image synthesis~\cite{odena2017conditional, wang2018high}.

In more recent times, text-to-image generation models have attracted interest following the introduction of Diffusion models~\cite{dhariwal2021diffusion, nichol2021glide}. Most of the recent Diffusion based image synthesis models, including Stable Diffusion~\cite{rombach2022high}, SDXL~\cite{podell2023sdxl}, DALL-E~\cite{ramesh2022hierarchical}, Imagen~\cite{saharia2022photorealistic} have demonstrated the ability to produce high quality images. Diffusion models also demonstrate the ability to generate images spanning a diverse range of categories and scenes as compared to GANs.

With the widespread availability of powerful open-source fake image synthesis models, the necessity to develop models capable of detecting fake images has become more crucial than ever before. Numerous previously proposed deepfake image detection methods opted to learn a deep neural network classifier capable of classifying $real$ vs $fake$ images originating from the same generative model~\cite{rossler2019faceforensics++}. However, studies suggest that such classifiers do not generalize well onto detecting fake images coming from other distribution than the training one~\cite{zhu2021face, khan2023deepfake}. 

\citeauthor{wang2020cnn}~\cite{wang2020cnn} proposed a simple yet effective solution to the challenge of detecting images generated by GANs. By training a well-known CNN architecture, ResNet-50~\cite{he2016deep}, on a single GAN-generated dataset (ProGAN~\cite{karras2017progressive}), along with augmentations like JPEG compression and blurring, they significantly improved the model's robustness. This approach performed well even on images generated by different GAN models. Building on this, ~\citeauthor{gragnaniello2021gan}~\cite{gragnaniello2021gan} modified ResNet-50 for GAN image detection. They avoided down-sampling in initial layers in order to preserve high frequency GAN realted fingerprints, and applied intense augmentations during training, outperforming previous method~\cite{wang2020cnn}. \citeauthor{corvi2023detection}~\cite{corvi2023detection} extended work proposed in~\cite{gragnaniello2021gan}, training the same modified ResNet-50 on the dataset from~\cite{wang2020cnn}. They found their model excelled on GAN images but struggled with Diffusion models. However, training on images from LDMs~\cite{rombach2022high} yielded success on Diffusion-generated images but not on GAN ones. In a recent study, \citeauthor{ojha2023towards}~\cite{ojha2023towards} noted that previous techniques~\cite{wang2020cnn} fail on Diffusion model-generated images when initially trained on images generated by GAN models. They utilized a fixed CLIP encoder to train a linear classifier on CLIP features, achieving SOTA results for both GAN and Diffusion model-generated images by just training their model on GAN generated images same as~\cite{wang2020cnn, gragnaniello2021gan}.


\begin{figure*}[t!]
  \centering
  \includegraphics[width=0.95\linewidth]{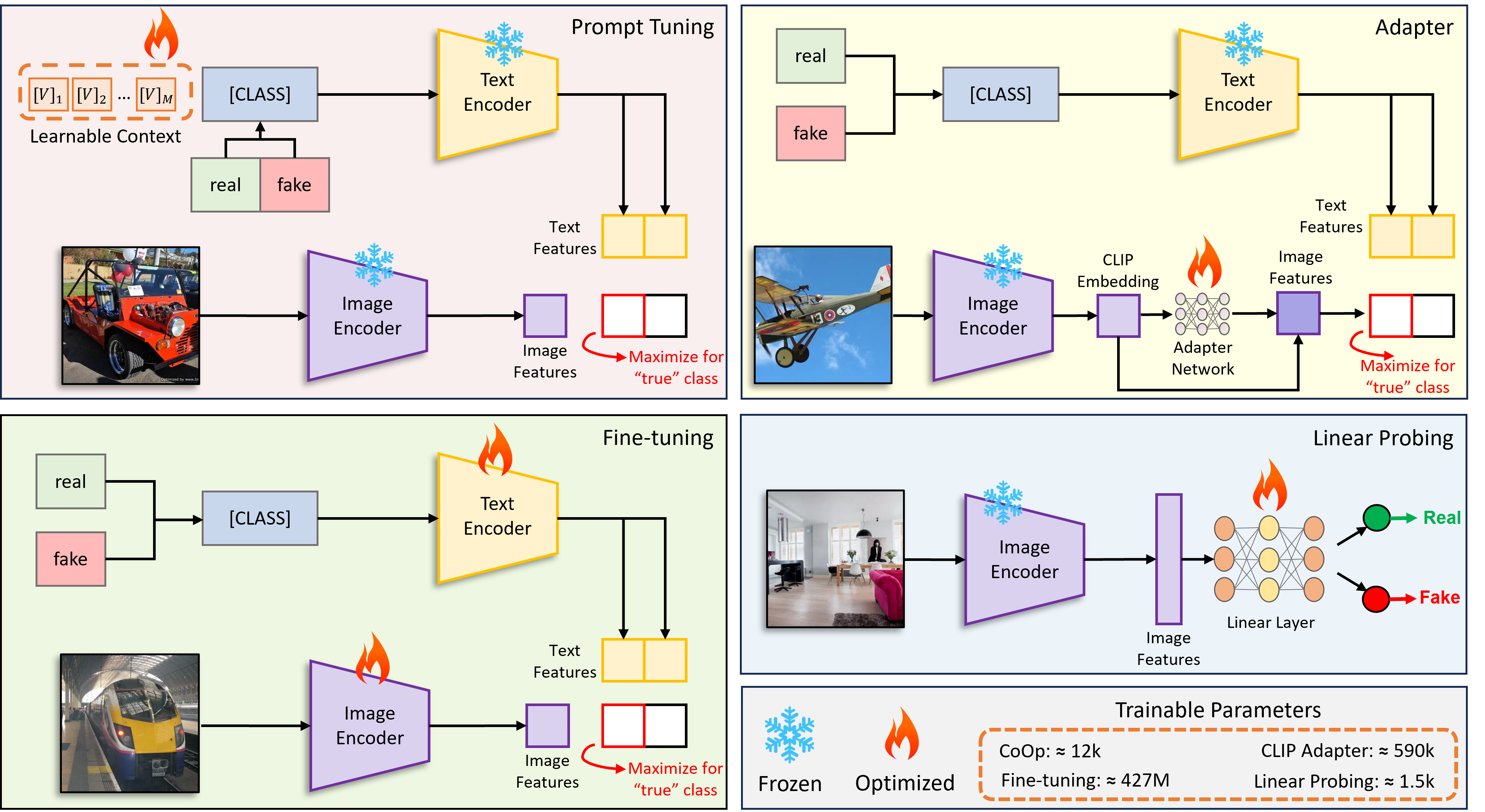}
\caption{In this figure, we present four distinct transfer learning strategies that are explored for $real/fake$ image classification. At bottom right we list the number of trainable parameters for each approach.}
\label{fig:main}
\end{figure*}

\section{Methodology}
\label{sec:methodology}
\subsection{Background}
The ultimate objective of a deepfake detection system is to determine if any given image is (a) authentic: captured using a camera, or (b) fake: synthesized using a generative model (GAN or Diffusion). In this section, we outline the methodologies examined in this study for training our detection model, along with the datasets used to train and evaluate our model. However, we begin by first presenting the baseline~\cite{wang2018high} and current SOTA~\cite{ojha2023towards} approaches proposed recently to address this task. These studies effectively leads us towards our proposed solution.

\citeauthor{wang2020cnn} in \cite{wang2020cnn} trained a ResNet-50~\cite{he2016deep} using cross-entropy loss to perform binary classification between $real$ and $fake$ images, using data they generated using the ProGAN model~\cite{karras2017progressive} after training it on 20 different object categories taken originally from LSUN~\cite{yu2015lsun}. For each of the 20 object categories, the authors generated 18k synthetic images, totaling up to 360k $fake$ images. They incorporated $real$ images from the LSUN dataset, amounting to 18k $real$ images for each of the 20 object categories. Consequently, their training dataset contained 720k $real$ and $fake$ images. They demonstrated through comprehensive evaluation that a simple CNN, when trained with meticulous data augmentation techniques like compression and blurring, exhibits effective generalization for deepfake detection on previously unseen data. They evaluated their trained model on images synthesized by various different GAN models showing excellent results.

Following this, \citeauthor{ojha2023towards}~\cite{ojha2023towards} found that the work in~\cite{wang2020cnn} was not performing as expected when tested on images synthesized by Diffusion models. For instance, on images generated by models like Latent~\cite{rombach2022high} and Guided~\cite{dhariwal2021diffusion} Diffusion models, the detection model's classification accuracy experiences a significant decline, reaching close to chance performance. This implies that during training, the model emphasizes solely on detecting the presence or absence of model specific artifacts in an image, while overlooking other distinguishing features between $real$ and $fake$ images. As a consequence, the resulting model becomes biased towards a single class ($real$ in this case), leading to the misclassification of $fake$ images from a Diffusion model without GAN-specific artifacts as $real$.

To tackle this issue, the authors suggested that the classification process should occur in a feature space that has not been solely learned to discriminate between $real$ and $fake$ images. This approach was aimed at preventing bias towards recognizing specific artifacts from one class (Real, GAN, or Diffusion) disproportionately better than the other~\cite{ojha2023towards}. Additionally, the selected feature space must capture a wide range of images, ensuring a robust fake image detector that works reliably across various categories such as outdoor scenes, objects, faces and beyond. The authors identified that CLIP's~\cite{radford2021learning} feature space possesses these desirable qualities – it was not initially trained for $real$ vs $fake$ classification, and has been exposed to a variety of images representing diverse objects and scenes. 

To validate their hypothesis, the authors used CLIP's image encoder (ViT-Large) as a feature extractor, and trained a simple linear model on top. They used the same dataset as in~\cite{wang2020cnn} training. The obtained results supported their hypothesis: their simple approach achieved state-of-the-art performance on previously unseen images from both GAN and Diffusion models~\cite{ojha2023towards}. While~\cite{ojha2023towards} achieves excellent results on most datasets, it still seems to struggle on some datasets, including Guided Diffusion~\cite{dhariwal2021diffusion}, LDM~\cite{rombach2022high}, Deepfakes~\cite{rossler2019faceforensics++}, FaceSwap~\cite{rossler2019faceforensics++}, and Commercial generators such as DALL-E 3~\footnote{https://openai.com/dall-e-3}, Adobe Firefly~\footnote{https://www.adobe.com/products/firefly.html} and Midjourney~\footnote{https://www.midjourney.com/}.


\subsection{Transfer Learning}

When applied to adapt vision-language models for downstream vision tasks, linear probing faces a significant drawback as it completely overlooks the language component. As noted in~\cite{zhou2022learning}, a linear layer trained on visual features serves as a static set of weights exclusively representing visual concepts. Consequently, the semantics embedded in texts remain largely unexplored, and irrelevant during this process. This limitation is exemplified in \cite{ojha2023towards}, where only the visual component of CLIP is utilized for deepfake detection, while completely neglecting the text encoder. We believe that leveraging both the visual and text encoders of CLIP~\cite{radford2021learning} can lead to an improved strategy for $real$ vs $fake$ classification. 

Based on this insight, we propose leveraging CoOp~\cite{zhou2022learning}, a Prompt Tuning strategy as our central approach to adapt CLIP~\cite{radford2021learning} for deepfake detection. Prompt tuning is particularly appealing as it integrates both the visual and language aspects of CLIP. To ensure a fair assessment of the robustness of various transfer learning strategies, we incorporate three additional methods, in addition to Prompt Tuning for this task, including (1) Linear Probing, (2) Full Fine-tuning and (3) training an Adapter Network~\cite{gao2023clip}. A concise overview of each employed transfer learning strategy is presented in the following sections.

\subsubsection{\textbf{Linear Probing:}}
Linear probing, a well-known transfer learning strategy, involves fine-tuning a linear classifier on top of a frozen model (CLIP in our case). We follow the same approach as employed by~\citeauthor{ojha2023towards}~\cite{ojha2023towards}, i.e., we discard CLIP's text encoder while freezing its image encoder. We then train a single linear layer for classification on the frozen CLIP's image features, mapping the penultimate image features to logits for class predictions using the Sigmoid activation function. The optimization takes place using the binary cross entropy loss. We illustrate linear probing strategy in Figure~\ref{fig:main}.




\subsubsection{\textbf{Fine-tuning:}}
Fine-tuning in this context means training the whole CLIP model (ViT-Large) again on the downstream dataset, which in our case is the ProGAN dataset which was also used by~\cite{wang2020cnn} and~\cite{ojha2023towards}. Full fine-tuning requires significantly more compute resources, data, and training time since the entire model is retrained.  Additionally, as model size increases, this strategy demonstrates instability and inefficiency~\cite{kim2021adapt}. During the training of our models, we encountered this issue, and mitigated it by utilizing an extremely small learning rate, $1 \times 10^{-6}$. To fine-tune our model, we adhere to the procedure outlined in the pre-training of CLIP~\cite{radford2021learning}. However, we introduce a modification: rather than utilizing entire text captions for each image, we provide only single-word captions, specifically either $real$ or $fake$. A typical Fine-tuning pipeline for adapting CLIP is illustrated in Figure~\ref{fig:main}.


\begin{table}[t!]
  \centering
  \small
  \caption{This table showcases the statistics of the test datasets. Certain datasets include their own collection of $real$ images. However, for datasets that lack their own $real$ images, we utilize LAION's~\cite{schuhmann2021laion} images instead. }
  \resizebox{0.85\linewidth}{!}{%
    \begin{tabular}{lccc}
    \toprule
    \multirow{2}{*}{Generator} & \multicolumn{1}{p{4.215em}}{\centering Num. $real$/$fake$} & \multicolumn{1}{p{4.215em}}{\centering Real Data Source} & \multicolumn{1}{p{5.215em}}{\centering Image Resolution} \\
    \midrule
    ProGAN~\cite{karras2017progressive}  & 4k / 4k & LSUN & 256 x 256 \\
    BigGAN~\cite{brock2018large}  & 2k / 2k  & ImageNet & 256 x 256 \\
    CycleGAN~\cite{zhu2017unpaired}  & 1k / 1k & Various & 256 x 256 \\
    EG3D~\cite{chan2022efficient}  & 1k / 1k & LAION & 512 x 512 \\
    GauGAN~\cite{park2019semantic}  & 5k / 5k & COCO & 256 x 256 \\
    StarGAN~\cite{choi2018stargan}  & 2k / 2k & CelebA & 256 x 256 \\
    StyleGAN~\cite{karras2019style} & 1k / 1k & LSUN & 256 x 256 \\
    StyleGAN2~\cite{karras2020analyzing}  & 1k / 1k & Various & $\approx$ 256 x 256 \\
    StyleGAN3~\cite{karras2021alias}  & $\approx$ 1k / 1k & Various & 512 x 512 \\
    Taming-T~\cite{esser2021taming}  & 1k / 1k & LAION & 256 x 256 \\
    DALL-E (mini)~\cite{DaymaDallemini}  & 1k / 1k & LAION & 256 x 256  \\
    Glide~\cite{nichol2021glide}  & 1k / 1k & LAION & 256 x 256 \\
    Guided~\cite{guidedDiff}  & 1k / 1k & LAION & 256 x 256 \\
    LDM~\cite{rombach2022high}  & 1k / 1k & LAION & 256 x 256 \\
    Stable Diff.~\cite{rombach2022high}  & 1k / 1k & LAION & 512 x 512 \\
    SDXL~\cite{podell2023sdxl}  & 1k / 1k & LAION & 1024 x 1024 \\
    Deepfakes~\cite{rossler2019faceforensics++}  & $\approx$ 2.7k / 2.7k  & YouTube & $\approx$ 256 x 256 \\
    FaceSwap~\cite{rossler2019faceforensics++}  & 2.8k / 2.8k & YouTube & $\approx$ 256 x 256 \\
    Midjourney-V5  & 1k / 1k & LAION & Various \\
    Adobe Firefly  & 1k / 1k & LAION & Various \\
    DALL-E 3  & 1k / 1k & LAION & Various \\
    
    \bottomrule
    \end{tabular}%
    }
  \label{tab:datasetstats}%
\end{table}

\begin{table*}[htbp]
  \centering
  \caption{Generalization performance. This table presents the average precision (AP) of different methods for distinguishing $real$ and $fake$ images. The studied adaptation approaches demonstrate significant improvements over the previous baselines and SOTA.}
  \resizebox{\linewidth}{!}{%
    \begin{tabular}{lcccccccccccccccccccc}
    \toprule
    \multirow{2}{*}{Method} & \multirow{2}{*}{Variant} & \multicolumn{10}{c}{Generative Adversarial Networks} & \multirow{2}{*}{DALL-E} & \multicolumn{5}{c}{Denoising Diffusion Models} & \multicolumn{2}{c}{FF++} & \multirow{2}{*}{mAP} \\
    \cmidrule(lr){3-12} \cmidrule(lr){14-18} \cmidrule(lr){19-20}
          &  & \multicolumn{1}{p{3.215em}}{\centering Pro GAN} & \multicolumn{1}{p{3.215em}}{\centering Big GAN} & \multicolumn{1}{p{3.215em}}{\centering Cycle GAN} & {\multirow{1}[4]{*} {\centering EG3D}} & \multicolumn{1}{p{3.215em}}{\centering Gau GAN} & \multicolumn{1}{p{3.215em}}{\centering Star GAN} & \multicolumn{1}{p{3.215em}}{\centering Style GAN} & \multicolumn{1}{p{3.215em}}{\centering Style GAN-2} & \multicolumn{1}{p{3.215em}}{\centering Style GAN-3} & {\multirow{1}[4]{*} {\centering Taming-T}} & & {\multirow{1}[4]{*} {\centering Glide}} & {\multirow{1}[4]{*} {\centering Guided}} & {\multirow{1}[4]{*} {\centering LDM}} & {\multirow{1}[4]{*} {\centering SD}} & {\multirow{1}[4]{*} {\centering SDXL}} & \multicolumn{1}{p{3.215em}}{\centering Deep Fakes} & \multicolumn{1}{p{3.215em}}{\centering Face Swap} &  \\
    \midrule
        \multirow{2}{*}{\begin{tabular}[l]{@{}l@{}}\citeauthor{wang2020cnn}\\(CVPR'20)\end{tabular}}  & Blur+JPEG (0.1) & \textbf{100.00} & 83.04 & 90.09 & 95.58 & 88.94 & 97.18 & 99.27 & 96.43 & 98.63 & 73.90 & 67.47 & 81.02 & 83.10 & 68.61 & 64.33 & 72.27 & 75.88 & 50.78 & 81.18 \\

        & Blur+JPEG (0.5) & \textbf{100.00} & 82.63 & 94.71 & 55.32 & 96.62 & 93.88 & 93.25 & 88.64 & 85.33 & 59.78 & 60.92 & 69.75 & 65.11 & 60.24 & 52.14 & 65.92 & 64.33 & 49.76 & 72.65 \\

    \midrule
        \multirow{2}{*}{\begin{tabular}[l]{@{}l@{}}Gragn. et al.\\(ICME'21)\end{tabular}}  & \multirow{2}{*}{\begin{tabular}[l]{@{}c@{}}ResNet-50\\No Downsample\end{tabular}} & \multirow{2}{*}{\begin{tabular}[l]{@{}l@{}}\textbf{100.00}\end{tabular}} & \multirow{2}{*}{\begin{tabular}[l]{@{}l@{}}97.57\end{tabular}} & \multirow{2}{*}{\begin{tabular}[l]{@{}l@{}}97.63\end{tabular}} & \multirow{2}{*}{\begin{tabular}[l]{@{}l@{}}99.95\end{tabular}} & \multirow{2}{*}{\begin{tabular}[l]{@{}l@{}}98.36\end{tabular}} & \multirow{2}{*}{\begin{tabular}[l]{@{}l@{}}99.99\end{tabular}} & \multirow{2}{*}{\begin{tabular}[l]{@{}l@{}}\textbf{100.00}\end{tabular}} & \multirow{2}{*}{\begin{tabular}[l]{@{}l@{}}\textbf{99.98}\end{tabular}} & \multirow{2}{*}{\begin{tabular}[l]{@{}l@{}}\textbf{100.00}\end{tabular}} & \multirow{2}{*}{\begin{tabular}[l]{@{}l@{}}95.31\end{tabular}} & \multirow{2}{*}{\begin{tabular}[l]{@{}l@{}}91.32\end{tabular}} & \multirow{2}{*}{\begin{tabular}[l]{@{}l@{}}94.08\end{tabular}} & \multirow{2}{*}{\begin{tabular}[l]{@{}l@{}}93.81\end{tabular}} & \multirow{2}{*}{\begin{tabular}[l]{@{}l@{}}92.33\end{tabular}} & \multirow{2}{*}{\begin{tabular}[l]{@{}l@{}}91.75\end{tabular}} & \multirow{2}{*}{\begin{tabular}[l]{@{}l@{}}90.93\end{tabular}} & \multirow{2}{*}{\begin{tabular}[l]{@{}l@{}}\textbf{95.90}\end{tabular}} & \multirow{2}{*}{\begin{tabular}[l]{@{}l@{}}61.54\end{tabular}} & \multirow{2}{*}{\begin{tabular}[l]{@{}l@{}}94.24\end{tabular}} \\
        \\
    \midrule
        \multirow{2}{*}{\begin{tabular}[l]{@{}l@{}}\citeauthor{corvi2023detection}\\(ICASSP'23)\end{tabular}}  & \multirow{1}{*}{\begin{tabular}[l]{@{}l@{}}ProGAN/LSUN\end{tabular}} & \multirow{1}{*}{\begin{tabular}[l]{@{}l@{}}\textbf{100.00}\end{tabular}} & \multirow{1}{*}{\begin{tabular}[l]{@{}l@{}}\textbf{99.66}\end{tabular}} & \multirow{1}{*}{\begin{tabular}[l]{@{}l@{}}97.94\end{tabular}} & \multirow{1}{*}{\begin{tabular}[l]{@{}l@{}}99.92\end{tabular}} & \multirow{1}{*}{\begin{tabular}[l]{@{}l@{}}99.74\end{tabular}} & \multirow{1}{*}{\begin{tabular}[l]{@{}l@{}}99.95\end{tabular}} & \multirow{1}{*}{\begin{tabular}[l]{@{}l@{}}\textbf{100.00}\end{tabular}} & \multirow{1}{*}{\begin{tabular}[l]{@{}l@{}}\textbf{99.96}\end{tabular}} & \multirow{1}{*}{\begin{tabular}[l]{@{}l@{}}99.93\end{tabular}} & \multirow{1}{*}{\begin{tabular}[l]{@{}l@{}}94.34\end{tabular}} & \multirow{1}{*}{\begin{tabular}[l]{@{}l@{}}95.45\end{tabular}} & \multirow{1}{*}{\begin{tabular}[l]{@{}l@{}}89.51\end{tabular}} & \multirow{1}{*}{\begin{tabular}[l]{@{}l@{}}79.30\end{tabular}} & \multirow{1}{*}{\begin{tabular}[l]{@{}l@{}}88.26\end{tabular}} & \multirow{1}{*}{\begin{tabular}[l]{@{}l@{}}87.01\end{tabular}} & \multirow{1}{*}{\begin{tabular}[l]{@{}l@{}}74.90\end{tabular}} & \multirow{1}{*}{\begin{tabular}[l]{@{}l@{}}95.52\end{tabular}} & \multirow{1}{*}{\begin{tabular}[l]{@{}l@{}}56.58\end{tabular}} & \multirow{1}{*}{\begin{tabular}[l]{@{}l@{}}91.52\end{tabular}} \\
        
        & \multirow{1}{*}{\begin{tabular}[l]{@{}l@{}}Latent/LSUN\end{tabular}} & \multirow{1}{*}{\begin{tabular}[l]{@{}l@{}}91.83\end{tabular}} & \multirow{1}{*}{\begin{tabular}[l]{@{}l@{}}74.25\end{tabular}} & \multirow{1}{*}{\begin{tabular}[l]{@{}l@{}}49.05\end{tabular}} & \multirow{1}{*}{\begin{tabular}[l]{@{}l@{}}42.87\end{tabular}} & \multirow{1}{*}{\begin{tabular}[l]{@{}l@{}}89.14\end{tabular}} & \multirow{1}{*}{\begin{tabular}[l]{@{}l@{}}50.19\end{tabular}} & \multirow{1}{*}{\begin{tabular}[l]{@{}l@{}}73.25\end{tabular}} & \multirow{1}{*}{\begin{tabular}[l]{@{}l@{}}74.73\end{tabular}} & \multirow{1}{*}{\begin{tabular}[l]{@{}l@{}}70.20\end{tabular}} & \multirow{1}{*}{\begin{tabular}[l]{@{}l@{}}95.21\end{tabular}} & \multirow{1}{*}{\begin{tabular}[l]{@{}l@{}}98.15\end{tabular}} & \multirow{1}{*}{\begin{tabular}[l]{@{}l@{}}87.35\end{tabular}} & \multirow{1}{*}{\begin{tabular}[l]{@{}l@{}}59.17\end{tabular}} & \multirow{1}{*}{\begin{tabular}[l]{@{}l@{}}\textbf{100.00}\end{tabular}} & \multirow{1}{*}{\begin{tabular}[l]{@{}l@{}}\textbf{100.00}\end{tabular}} & \multirow{1}{*}{\begin{tabular}[l]{@{}l@{}}99.23\end{tabular}} & \multirow{1}{*}{\begin{tabular}[l]{@{}l@{}}83.70\end{tabular}} & \multirow{1}{*}{\begin{tabular}[l]{@{}l@{}}45.52\end{tabular}} & \multirow{1}{*}{\begin{tabular}[l]{@{}l@{}}79.93\end{tabular}}
        \\
    \midrule
        \multirow{2}{*}{\begin{tabular}[l]{@{}l@{}}\citeauthor{ojha2023towards}\\(CVPR'23)\end{tabular}} & \multirow{2}{*}{\begin{tabular}[l]{@{}c@{}}CLIP\\Linear Probing\end{tabular}} & \multirow{2}{*}{\begin{tabular}[l]{@{}l@{}}99.99\end{tabular}} & \multirow{2}{*}{\begin{tabular}[l]{@{}l@{}}98.73\end{tabular}} & \multirow{2}{*}{\begin{tabular}[l]{@{}l@{}}98.92\end{tabular}} & \multirow{2}{*}{\begin{tabular}[l]{@{}l@{}}79.58\end{tabular}} & \multirow{2}{*}{\begin{tabular}[l]{@{}l@{}}99.74\end{tabular}} & \multirow{2}{*}{\begin{tabular}[l]{@{}l@{}}96.06\end{tabular}} & \multirow{2}{*}{\begin{tabular}[l]{@{}l@{}}95.73\end{tabular}} & \multirow{2}{*}{\begin{tabular}[l]{@{}l@{}}95.81\end{tabular}} & \multirow{2}{*}{\begin{tabular}[l]{@{}l@{}}92.21\end{tabular}} & \multirow{2}{*}{\begin{tabular}[l]{@{}l@{}}97.12\end{tabular}}  & \multirow{2}{*}{\begin{tabular}[l]{@{}l@{}}96.84\end{tabular}} & \multirow{2}{*}{\begin{tabular}[l]{@{}l@{}}93.85\end{tabular}} & \multirow{2}{*}{\begin{tabular}[l]{@{}l@{}}92.09\end{tabular}} & \multirow{2}{*}{\begin{tabular}[l]{@{}l@{}}95.71\end{tabular}} & \multirow{2}{*}{\begin{tabular}[l]{@{}l@{}}93.58\end{tabular}} & \multirow{2}{*}{\begin{tabular}[l]{@{}l@{}}88.55\end{tabular}} & \multirow{2}{*}{\begin{tabular}[l]{@{}l@{}}77.48\end{tabular}} & \multirow{2}{*}{\begin{tabular}[l]{@{}l@{}}75.87\end{tabular}} & \multirow{2}{*}{\begin{tabular}[l]{@{}l@{}}93.05\end{tabular}}  \\

        \\

    \midrule
        
        \multirow{4}{*}{Ours} & 
        Linear Probing & 99.91 & 97.77 & 98.53 & 99.48 & 99.69 & 99.00 & 95.53 & 94.98 & 99.54 & 97.74 & 95.65 & 97.75 & 92.14 & 95.94 & 92.24 & 94.99 & 80.07 & 76.58 & 95.22 \\
        & Fine Tuning & \textbf{100.00} & 98.65 & 99.00 & \textbf{99.97} & 98.12 & \textbf{100.00} & 99.61 & 99.48 & \textbf{100.00} & 98.38 & 98.15 & 96.23 & 97.40 & 98.79 & 97.53  & \textbf{99.52} & 87.42 & 60.22 & 96.29 \\
        & Adapter & \textbf{100.00} & \textbf{99.58} & \textbf{99.97} & 99.50 & \textbf{99.98} & 99.98 & 99.44 & 98.80 & 99.83 & 99.27  & 98.60 & 99.26 & 96.16 & 97.76 & 91.90 & 92.32 & 91.37 & 82.11 & 97.27 \\
        & Prompt Tuning & \textbf{100.00} & 99.42 & 99.92 & 99.51 & 99.95 & 99.97 & 99.52 & 98.62 & 99.68 & \textbf{99.54} & \textbf{98.89} & \textbf{99.32} & \textbf{97.41} & 97.91 & 96.23 & 96.42 & 92.59 & \textbf{88.01} & \textbf{98.06} \\
    
    \bottomrule
    \end{tabular}%
    }
  \label{tab:allAP}%
\end{table*}%

\begin{table*}[htbp]
  \centering
  \caption{Generalization performance. This table compares the accuracy (Acc) scores attained by our proposed techniques with various previous studies. The proposed CLIP adaptation strategies show noteworthy performance gains compared to previous baselines and SOTA techniques.}
  \resizebox{\linewidth}{!}{%
    \begin{tabular}{lcccccccccccccccccccc}
    \toprule
    \multirow{2}{*}{Method} & \multirow{2}{*}{Variant} & \multicolumn{10}{c}{Generative Adversarial Networks} & \multirow{2}{*}{DALL-E} & \multicolumn{5}{c}{Denoising Diffusion Models} & \multicolumn{2}{c}{FF++} & \multirow{2}{*}{\begin{tabular}[c]{@{}c@{}}Avg.\\Acc\end{tabular}} \\
    \cmidrule(lr){3-12} \cmidrule(lr){14-18} \cmidrule(lr){19-20}
          &  & \multicolumn{1}{p{3.215em}}{\centering Pro GAN} & \multicolumn{1}{p{3.215em}}{\centering Big GAN} & \multicolumn{1}{p{3.215em}}{\centering Cycle GAN} & {\multirow{1}[4]{*} {\centering EG3D}} & \multicolumn{1}{p{3.215em}}{\centering Gau GAN} & \multicolumn{1}{p{3.215em}}{\centering Star GAN} & \multicolumn{1}{p{3.215em}}{\centering Style GAN} & \multicolumn{1}{p{3.215em}}{\centering Style GAN-2} & \multicolumn{1}{p{3.215em}}{\centering Style GAN-3} & {\multirow{1}[4]{*} {\centering Taming-T}} & & {\multirow{1}[4]{*} {\centering Glide}} & {\multirow{1}[4]{*} {\centering Guided}} & {\multirow{1}[4]{*} {\centering LDM}} & {\multirow{1}[4]{*} {\centering SD}} & {\multirow{1}[4]{*} {\centering SDXL}} & \multicolumn{1}{p{3.215em}}{\centering Deep Fakes} & \multicolumn{1}{p{3.215em}}{\centering Face Swap} &  \\
    \midrule
        \multirow{2}{*}{\begin{tabular}[l]{@{}l@{}}\citeauthor{wang2020cnn}\\(CVPR'20)\end{tabular}}  & Blur+JPEG (0.1) & 99.90 & 67.65 & 79.50 & 72.65 & 76.63 & 89.72 & 82.10 & 77.05 & 80.68 & 56.45 & 55.05 & 61.15 & 62.90 & 54.03 & 52.50 & 53.40 & 52.67 & 49.68 & 66.09 \\

        & Blur+JPEG (0.5) & 99.65 & 58.13 & 77.80 & 50.30 & 75.56 & 79.99 & 69.80 & 62.30 & 53.42 & 51.05 & 51.90 & 54.33 & 52.35 & 51.35 & 50.15 & 51.00  & 51.46 & 50.02 & 59.18 \\

    \midrule
        \multirow{2}{*}{\begin{tabular}[l]{@{}l@{}}Gragn. et al.\\(ICME'21)\end{tabular}} & \multirow{2}{*}{\begin{tabular}[l]{@{}c@{}}ResNet-50\\No Downsample\end{tabular}} & \multirow{2}{*}{\begin{tabular}[l]{@{}l@{}}\textbf{100.00}\end{tabular}} & \multirow{2}{*}{\begin{tabular}[l]{@{}l@{}}93.27\end{tabular}} & \multirow{2}{*}{\begin{tabular}[l]{@{}l@{}}91.75\end{tabular}} & \multirow{2}{*}{\begin{tabular}[l]{@{}l@{}}97.55\end{tabular}} & \multirow{2}{*}{\begin{tabular}[l]{@{}l@{}}94.13\end{tabular}} & \multirow{2}{*}{\begin{tabular}[l]{@{}l@{}}99.65\end{tabular}} & \multirow{2}{*}{\begin{tabular}[l]{@{}l@{}}97.25\end{tabular}} & \multirow{2}{*}{\begin{tabular}[l]{@{}l@{}}89.75\end{tabular}} & \multirow{2}{*}{\begin{tabular}[l]{@{}l@{}}97.47\end{tabular}} & \multirow{2}{*}{\begin{tabular}[l]{@{}l@{}}67.45\end{tabular}} & \multirow{2}{*}{\begin{tabular}[l]{@{}l@{}}60.65\end{tabular}} & \multirow{2}{*}{\begin{tabular}[l]{@{}l@{}}69.38\end{tabular}} & \multirow{2}{*}{\begin{tabular}[l]{@{}l@{}}67.30\end{tabular}} & \multirow{2}{*}{\begin{tabular}[l]{@{}l@{}}62.33\end{tabular}} & \multirow{2}{*}{\begin{tabular}[l]{@{}l@{}}59.70\end{tabular}} & \multirow{2}{*}{\begin{tabular}[l]{@{}l@{}}57.75\end{tabular}} & \multirow{2}{*}{\begin{tabular}[l]{@{}l@{}}65.31\end{tabular}} & \multirow{2}{*}{\begin{tabular}[l]{@{}l@{}}50.02\end{tabular}} & \multirow{2}{*}{\begin{tabular}[l]{@{}l@{}}76.59\end{tabular}} \\

        \\
    \midrule
        \multirow{2}{*}{\begin{tabular}[l]{@{}l@{}}\citeauthor{corvi2023detection}\\(ICASSP'23)\end{tabular}} & \multirow{1}{*}{\begin{tabular}[l]{@{}l@{}}ProGAN/LSUN\end{tabular}} & \multirow{1}{*}{\begin{tabular}[l]{@{}l@{}}\textbf{100.00}\end{tabular}} & \multirow{1}{*}{\begin{tabular}[l]{@{}l@{}}\textbf{95.85}\end{tabular}} & \multirow{1}{*}{\begin{tabular}[l]{@{}l@{}}90.35\end{tabular}} & \multirow{1}{*}{\begin{tabular}[l]{@{}l@{}}\textbf{98.40}\end{tabular}} & \multirow{1}{*}{\begin{tabular}[l]{@{}l@{}}92.46\end{tabular}} & \multirow{1}{*}{\begin{tabular}[l]{@{}l@{}}99.00\end{tabular}} & \multirow{1}{*}{\begin{tabular}[l]{@{}l@{}}\textbf{97.65}\end{tabular}} & \multirow{1}{*}{\begin{tabular}[l]{@{}l@{}}84.90\end{tabular}} & \multirow{1}{*}{\begin{tabular}[l]{@{}l@{}}82.79\end{tabular}} & \multirow{1}{*}{\begin{tabular}[l]{@{}l@{}}65.30\end{tabular}} & \multirow{1}{*}{\begin{tabular}[l]{@{}l@{}}69.30\end{tabular}} & \multirow{1}{*}{\begin{tabular}[l]{@{}l@{}}58.98\end{tabular}} & \multirow{1}{*}{\begin{tabular}[l]{@{}l@{}}53.10\end{tabular}} & \multirow{1}{*}{\begin{tabular}[l]{@{}l@{}}58.83\end{tabular}} & \multirow{1}{*}{\begin{tabular}[l]{@{}l@{}}55.70\end{tabular}} & \multirow{1}{*}{\begin{tabular}[l]{@{}l@{}}52.10\end{tabular}} & \multirow{1}{*}{\begin{tabular}[l]{@{}l@{}}59.38\end{tabular}} & \multirow{1}{*}{\begin{tabular}[l]{@{}l@{}}50.11\end{tabular}} & \multirow{1}{*}{\begin{tabular}[l]{@{}l@{}}72.72\end{tabular}} \\

        & \multirow{1}{*}{\begin{tabular}[l]{@{}l@{}}Latent/LSUN\end{tabular}} & \multirow{1}{*}{\begin{tabular}[l]{@{}l@{}}50.94\end{tabular}} & \multirow{1}{*}{\begin{tabular}[l]{@{}l@{}}51.82\end{tabular}} & \multirow{1}{*}{\begin{tabular}[l]{@{}l@{}}46.20\end{tabular}} & \multirow{1}{*}{\begin{tabular}[l]{@{}l@{}}49.25\end{tabular}} & \multirow{1}{*}{\begin{tabular}[l]{@{}l@{}}50.86\end{tabular}} & \multirow{1}{*}{\begin{tabular}[l]{@{}l@{}}48.02\end{tabular}} & \multirow{1}{*}{\begin{tabular}[l]{@{}l@{}}59.40\end{tabular}} & \multirow{1}{*}{\begin{tabular}[l]{@{}l@{}}50.95\end{tabular}} & \multirow{1}{*}{\begin{tabular}[l]{@{}l@{}}50.05\end{tabular}} & \multirow{1}{*}{\begin{tabular}[l]{@{}l@{}}77.65\end{tabular}} & \multirow{1}{*}{\begin{tabular}[l]{@{}l@{}}87.00\end{tabular}} & \multirow{1}{*}{\begin{tabular}[l]{@{}l@{}}59.83\end{tabular}} & \multirow{1}{*}{\begin{tabular}[l]{@{}l@{}}50.95\end{tabular}} & \multirow{1}{*}{\begin{tabular}[l]{@{}l@{}}\textbf{99.25}\end{tabular}} & \multirow{1}{*}{\begin{tabular}[l]{@{}l@{}}\textbf{99.25}\end{tabular}} & \multirow{1}{*}{\begin{tabular}[l]{@{}l@{}}93.10\end{tabular}} & \multirow{1}{*}{\begin{tabular}[l]{@{}l@{}}69.87\end{tabular}} & \multirow{1}{*}{\begin{tabular}[l]{@{}l@{}}48.14\end{tabular}} & \multirow{1}{*}{\begin{tabular}[l]{@{}l@{}}66.40\end{tabular}}
        \\

    \midrule
        \multirow{2}{*}{\begin{tabular}[l]{@{}l@{}}\citeauthor{ojha2023towards}\\(CVPR'23)\end{tabular}} & \multirow{2}{*}{\begin{tabular}[l]{@{}c@{}}CLIP\\Linear Probing\end{tabular}} & \multirow{2}{*}{\begin{tabular}[l]{@{}l@{}}98.94\end{tabular}} & \multirow{2}{*}{\begin{tabular}[l]{@{}l@{}}94.48\end{tabular}} & \multirow{2}{*}{\begin{tabular}[l]{@{}l@{}}94.20\end{tabular}} & \multirow{2}{*}{\begin{tabular}[l]{@{}l@{}}57.75\end{tabular}} & \multirow{2}{*}{\begin{tabular}[l]{@{}l@{}}94.65\end{tabular}} & \multirow{2}{*}{\begin{tabular}[l]{@{}l@{}}87.49\end{tabular}} & \multirow{2}{*}{\begin{tabular}[l]{@{}l@{}}85.55\end{tabular}} & \multirow{2}{*}{\begin{tabular}[l]{@{}l@{}}83.40\end{tabular}} & \multirow{2}{*}{\begin{tabular}[l]{@{}l@{}}75.42\end{tabular}} & \multirow{2}{*}{\begin{tabular}[l]{@{}l@{}}89.45\end{tabular}} & \multirow{2}{*}{\begin{tabular}[l]{@{}l@{}}89.20\end{tabular}} & \multirow{2}{*}{\begin{tabular}[l]{@{}l@{}}82.15\end{tabular}} & \multirow{2}{*}{\begin{tabular}[l]{@{}l@{}}79.00\end{tabular}} & \multirow{2}{*}{\begin{tabular}[l]{@{}l@{}}87.80\end{tabular}} & \multirow{2}{*}{\begin{tabular}[l]{@{}l@{}}81.90\end{tabular}} & \multirow{2}{*}{\begin{tabular}[l]{@{}l@{}}74.15\end{tabular}} & \multirow{2}{*}{\begin{tabular}[l]{@{}l@{}}62.71\end{tabular}} & \multirow{2}{*}{\begin{tabular}[l]{@{}l@{}}64.30\end{tabular}} &  \multirow{2}{*}{\begin{tabular}[l]{@{}l@{}}82.84\end{tabular}} \\

        \\

    \midrule
        \multirow{4}{*}{Ours} & Linear Probing & 98.50 & 91.75 & 91.00 & 98.20 & 88.08 & 94.42 & 81.40 & 71.70 & 94.11 & 91.05 & 85.80 & 90.55 & 79.05 & 87.42 & 77.30 & 83.85  & 69.37 & 68.30 & 86.26 \\
          & Fine Tuning & 99.60 & 77.38 & 71.55 & \textbf{98.40} & 65.70 & \textbf{100.00} & 94.85 & \textbf{95.30} & \textbf{99.89} & 94.40 & \textbf{93.20} & 88.78 & \textbf{92.35} & 95.17 & 91.75 & \textbf{97.35}  & 76.46 & 52.11 & 88.74 \\
           & Adapter & 99.88 & 94.75 & \textbf{97.45} & 95.30 & \textbf{95.47} & 99.12 & 93.35 & 78.35 & 93.11 & 94.55 & 92.00 & \textbf{94.27} & 81.65 & 89.18 & 67.70 & 71.60  & 77.11 & 70.16 & 88.72 \\
           & Prompt Tuning & 99.83 & 93.80 & 95.60 & 93.50 & 93.43 & 99.15 & 95.25 & 82.95 & 93.11 & \textbf{94.95} & 91.50 & 92.88 & 84.3 & 88.16 & 76.45 & 77.80  & \textbf{78.45} & \textbf{74.66} &  \textbf{89.45}\\
    
    \bottomrule
    \end{tabular}%
    }
  \label{tab:allAcc}%
\end{table*}%

\subsubsection{\textbf{Prompt Tuning:}}

Initially introduced in the domain of natural language processing~\cite{liu2023pre}, Prompt Tuning is a relatively recent transfer learning strategy adopted by the computer vision community. This approach involves fine-tuning a pre-trained model like CLIP~\cite{radford2021learning} by learning randomly initialized prompts (textual~\cite{zhou2022learning} and/or visual~\cite{jia2022visual}) during training. The primary goal of Prompt Tuning is to adapt the model on specific downstream tasks by optimizing the prompts to align better with the target objectives.

In this study, we employ Context Optimization (CoOp), a transfer learning strategy introduced by~\citeauthor{zhou2022learning} in~\cite{zhou2022learning}, to fine-tune CLIP for the task of deepfake image detection. CoOp appends learnable vectors along with the context words\footnote{Context words refer to labels of any given dataset. In our case, the labels are $real$ and $fake$.} of a prompt. These learnable vectors can be either initialized with random values or pre-trained word embeddings~\cite{zhou2022learning}. During training the learnable vectors are optimized whereas both the text and vision encoders of CLIP are kept frozen.
\begin{equation}
t = [V]_1 [V]_2 \ldots [V]_M [CLASS]
\label{eq:1}
\end{equation}
Where each $[V]_m (m \in {1,...,M})$ is a vector with the same dimension as word embeddings, e.g., 768 for CLIP (ViT-Large)~\cite{zhou2022learning}. $M$ is a hyperparameter referring to the number of context tokens, i.e., $[V]_M$. We experiment with $M = [4, 8, 16, 24]$. $[CLASS]$ refers to class token of the dataset, e.g., $real$ and $fake$ in our case. Class token within each prompt $t_i$ is swapped with the corresponding word embedding vector of the $i$-th class name. The prompt $t$ is then fed through the text encoder, and optimized using cross-entropy loss during training. As evident from Eq.~\ref{eq:1}, the context tokens are added at the beginning of the class labels. While the CoOp paper explores various appending strategies, such as "end" and "middle", our findings indicate that appending context tokens at the "front" yields comparably better results. We show Prompt Tuning (CoOp) based CLIP training strategy in Figure~\ref{fig:main}.


\subsubsection{\textbf{Adapter Network:}}

Stepping away from Prompt Tuning, \citeauthor{gao2023clip} introduced a simple yet effective alternative approach for fine-tuning vision-language models using feature adapters~\cite{gao2023clip}. Specifically, the authors introduce CLIP-Adapter, an extra lightweight bottleneck layer which is optimized during training while the remainder of the CLIP model is kept frozen. Additionally, to remain robust against unseen data distributions, CLIP-Adapter integrates the original zero-shot visual or language embeddings with the corresponding fine-tuning feature embeddings through a ResNet styled residual connection~\cite{he2016deep}. This feature blending allows CLIP-Adapter to exploit both the knowledge stored in the original CLIP's feature space, and the newly acquired knowledge from the downstream training examples simultaneously. CLIP-Adapter can be applied to either the visual or language branch. In our study however, we only use Adapter Network with Vision branch, and leave the language branch as is. See Figure~\ref{fig:main} for reference.

\subsection{Generative Models Explored}
In this paper, we conduct an in-depth investigation into four distinct transfer learning approaches for deepfake detection. Our analysis is aimed at assessing the robustness of these approaches when coupled with pre-trained CLIP~\cite{radford2021learning} ViT-Large model for deepfake detection when exposed to unseen data coming from diverse deepfake generators including GANs and Diffusion models.

We follow the same protocols outlined by~\citeauthor{wang2020cnn}~\cite{wang2020cnn} and~\citeauthor{ojha2023towards}~\cite{ojha2023towards}, and train our models using data coming from just one generative model i.e., ProGAN~\cite{karras2017progressive}. However, for evaluation we incorporate an even broader spectrum of generative models in our analysis. This extension aims to align our evaluation more closely with real-world scenarios. In total, we assess our models across \textbf{21} distinct datasets, primarily categorized as GAN-based, Diffusion-based and commercial tools~\cite{corvi2023detection}. For detailed dataset statistics, please refer to Table~\ref{tab:datasetstats}.



Another minor fluctuation in evaluation protocol we follow is that \cite{ojha2023towards} employed three distinct configurations for image generation using Glide and LDMs, presenting their findings separately. In contrast, we include all images from Glide and LDM subsets in our analysis but display averaged results in our tables due to space constraints. 







\section{Experiments}
\label{sec:experiments}

In this section, we present performance scores achieved by CLIP ViT-Large~\cite{radford2021learning} when trained using four distinct transfer learning strategies: (1) Linear Probing, (2) Fine-tuning, (3) Adapter Network~\cite{gao2023clip} and (4) Prompt Tuning~\cite{zhou2022learning}. Additionally, we evaluate trained models released by~\cite{wang2020cnn, ojha2023towards, gragnaniello2021gan, corvi2023detection} on the same test set on which we evaluate our own models. Our aim is to determine if our chosen transfer learning strategies offer superior generalization compared to previous studies. In subsequent sections, besides assessing generalization capabilities, we conduct further experiments to assess performance of our models under various conditions, including smaller training set sizes, few-shot analysis, and robustness to post-processing operations.

\subsection{Generalization Performance}


We evaluate our model's performance by comparing it with four prior studies that aim to detect various types of deepfake images generated by different fake image generators. The initial study~\cite{wang2020cnn} in this field employed ResNet-50~\cite{he2016deep} as the classifier. They trained their models on $720k$ $real/fake$ images sourced from the ProGAN dataset which they generated for the sake of their study. They also employed image augmentations such as JPEG noise and Gaussian blurring, which made their models more robust towards post-processed images during evaluation. The second study~\cite{gragnaniello2021gan} also employs ResNet-50, but with a simple adjustment to the original architecture to better preserve the low-level forensic traces present inside images. The proposed modified model was also trained on the ProGAN dataset for $real/fake$ classification introduced in~\cite{wang2020cnn}. In~\cite{corvi2023detection} use the same modified ResNet-50~\cite{gragnaniello2021gan} but train it again on two different datasets, i.e., ProGAN/LSUN and LatentDiffusion/LSUN to better understand which generative model offers better generalization. The fourth study from~\citeauthor{ojha2023towards}~\cite{ojha2023towards} attained state-of-the-art performance. They utilized the CLIP ViT-Large model as a feature extractor, and subsequently trained a linear network on top of it for $real/fake$ classification.

In Tables~\ref{tab:allAP} and~\ref{tab:allAcc}, we compare our models' performance with that of~\cite{wang2020cnn, gragnaniello2021gan, corvi2023detection, ojha2023towards}. These studies~\cite{wang2020cnn, gragnaniello2021gan} demonstrate strong performance on GAN-generated images but show mediocre results on images from Diffusion-based and Commercial generators. Conversely, \cite{ojha2023towards} achieves good results on both GAN-based and Diffusion-based generators, although performance decreases on images from Commercial image generators (see Table~\ref{tab:robustnessacrossfamilies}) and FaceForensics++ dataset~\cite{rossler2019faceforensics++}, which utilizes an Auto-encoder based architecture for image synthesis.

Our four proposed CLIP adaptation approaches for deepfake detection demonstrate consistently better performance across all datasets as apparent from numbers in Tables~\ref{tab:allAP},~\ref{tab:allAcc} and~\ref{tab:robustnessacrossfamilies}. However, as seen in Tables~\ref{tab:allAP} and~\ref{tab:allAcc}, the Prompt Tuning strategy~\cite{zhou2022learning} notably outperforms other transfer learning strategies in terms of both mAP and average accuracy. Notably, Prompt Tuning optimizes only a fraction of parameters (\textbf{12k}) compared to the Adapter Network and full Fine-tuning approaches, which optimize a larger number of parameters. Overall, we surpass the previous SOTA~\cite{ojha2023towards} by \textbf{5.01\%} in mAP and \textbf{6.61\%} in average accuracy across images from \textbf{18} distinct synthetic image generators.


\begin{figure*}[t]
  \centering
  \includegraphics[width=\linewidth]{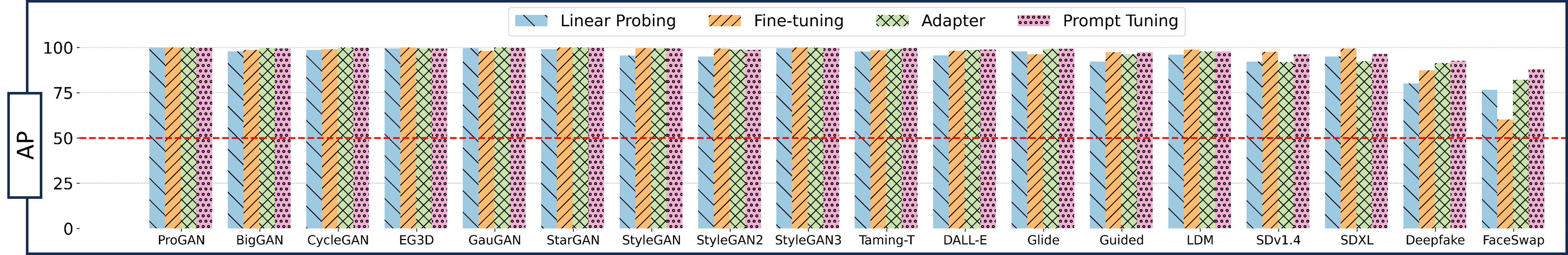}
\caption{Average precision (AP) score distribution of participating transfer learning strategies on the test set comprised of images sourced from 18 different datasets, as given in Tables~\ref{tab:allAP} and \ref{tab:allAcc}. The red dotted line represents chance performance.}
\label{fig:barAllAP}
\end{figure*}

\begin{figure*}[t]
  \centering
\includegraphics[width=\linewidth]{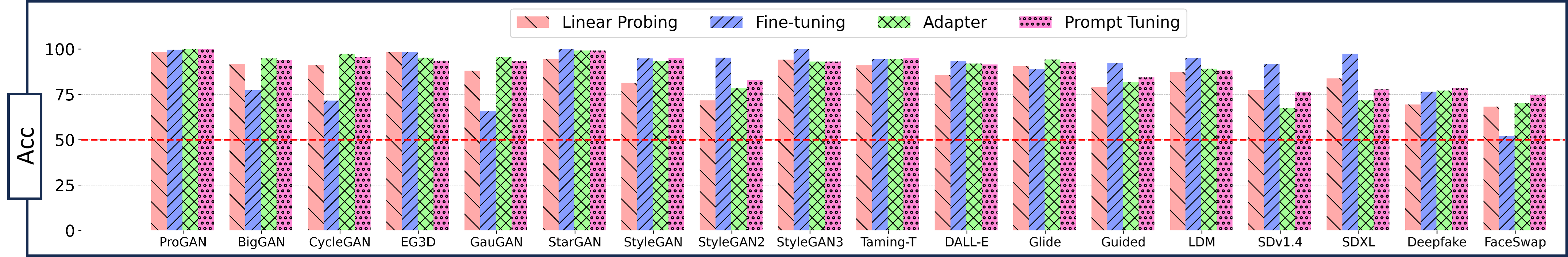}
\caption{Accuracy (Acc) scores achieved by participating transfer learning strategies on the test set comprised of images sourced from 18 different datasets, as given in Tables~\ref{tab:allAP} and \ref{tab:allAcc}. The red dotted line represents chance performance.}
\label{fig:barAllAcc}
\end{figure*}

\subsection{Effect of Transfer Learning Strategy}
In this section, we assess and compare the effectiveness of transfer learning strategies trained on images from ProGAN/LSUN datasets. Results are summarized in Tables~\ref{tab:allAP} and~\ref{tab:allAcc}. It is evident from the reported numbers that Prompt Tuning (CoOp) outperforms other strategies. Despite a modest margin, this is noteworthy as Prompt Tuning optimizes only a fraction of parameters ($\approx 12k$) compared to Linear Probing, Fine-tuning and Adapter Network, which optimize approximately ($\approx 1.5k$), ($\approx 427M$) and ($\approx 590k$) parameters respectively. Moreover, in few-shot experiments as shown in Table~\ref{tab:32shots} Prompt Tuning also outperforms other three strategies. However, in terms of robustness to post-processing operations, Linear Probing turns out to be best performing strategy. 


\subsection{Effect of Training Set Size}
We also conducted experiments with various training set sizes, and in this section, we report on the performance of participating transfer learning strategies when trained with reduced numbers of $real$ and $fake$ images. Using ProGAN's~\cite{karras2017progressive} data, we create four smaller datasets containing 20k, 40k, 60k and 80k images. As shown in Table~\ref{tab:trainsetsize}, we observe that while larger training datasets generally yield higher scores, the differences are not significant. Moreover, since the $fake$ images in the training data are generated by a GAN model (ProGAN~\cite{karras2017progressive}), the impact of training data size is less pronounced when evaluating models on other GAN models in the test set compared to Diffusion models, or Commercial tools. This analysis indicates that even with limited training resources, it is still possible to train robust detection models without a significant decline in generalization capabilities.

\subsection{Robustness to Post-processing Operations}
In real-world scenarios, images commonly undergo post-processing before being shared online, and research indicates that these operations significantly impact detection models' performance~\cite{wang2020cnn, ojha2023towards, cozzolino2023raising}. To assess how our models handle post-processing, following previous studies~\cite{wang2020cnn, ojha2023towards}, we evaluate them on images subjected to two types of operations: (1) JPEG compression and (2) Gaussian blurring.

\begin{figure*}[htbp]
  \centering
  \includegraphics[width=0.95\linewidth]{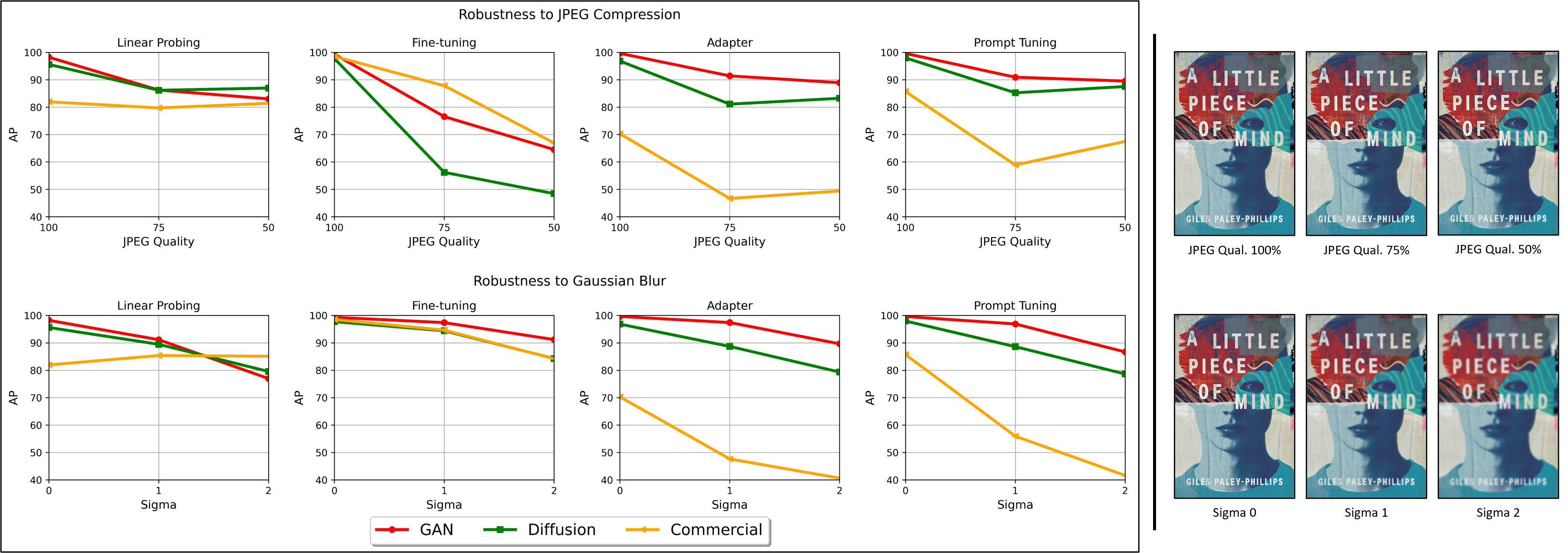}
\caption{
This figure shows how different transfer learning strategies cope with post-processing operations including JPEG compression and Gaussian blurring. Our models perform well with GAN and Diffusion images but struggle with those from commercial tools like DALL-E 3 and Adobe FireFly. Surprisingly, the Fine-tuned CLIP model is more robust against compressed images sampled using Commercial tools as compared to GAN-based and Diffusion-based images. Linear Probing achieves optimal performance across all three datasets.}
\label{fig:jpegrobustness}
\end{figure*}

To gauge the impact of JPEG compression, we tested two qualities: $75\%$ and $50\%$. For blurring effects, we used sigma values of $1$ and $2$. The performance results of our models are depicted in Figure~\ref{fig:jpegrobustness}. As expected, there is a decline in performance as sigma and compression values increase, though still acceptable considering our models weren't explicitly trained on compressed or blurred images. One thing we notice is that this decline is more pronounced for images generated by Commercial tools, except for fully Fine-tuned model. Linear Probing outperforms other adaptation strategies well across the three different generative model families.

\begin{table}[b!]
  \centering
  \caption{This table presents scores achieved by our models trained using samller sized datasets. Results are organized based on number of available training images: 20k, 40k, 60k and 80k. We keep equal amount of $real/fake$ images, e.g., for 20k subset, we have 10k $real$ and 10k $fake$ images.}
  \resizebox{\linewidth}{!}{%
    \begin{tabular}{lccccc}
    \toprule
    \multirow{3}{*}{Method} & \multirow{3}{*}{\begin{tabular}[c]{@{}c@{}}Num. Train\\Images\end{tabular}} & \multicolumn{3}{c}{Family of Generators} & \multirow{3}{*}{\begin{tabular}[c]{@{}c@{}}Average\\AP/Acc\end{tabular}} \\
    \cmidrule(lr){3-5} 
    & & \multicolumn{1}{p{5.215em}}{\centering GAN AP/Acc} & \multicolumn{1}{p{5.215em}}{\centering Diffusion AP/Acc} & \multicolumn{1}{p{6.215em}}{\centering Comm. Tools AP/Acc} & \\
    \midrule
        Linear Probing  & \multirow{4}{*}{20k} & 98.86 / \textbf{90.78} & \textbf{97.13} / \textbf{90.76} & \textbf{80.38} / \textbf{74.93} & \textbf{92.12} / \textbf{85.49}\\
        Fine-tuning  & & 95.68 / 78.10 & 86.79 / 69.39 & 71.45 / 63.78 & 84.64 / 70.43 \\
        Adapter  & & 98.57 / 89.17 & 93.51 / 83.81 & 62.01 / 53.60 & 84.70 / 75.53 \\
        Prompt Tuning  & & \textbf{98.95} / 90.42 & 96.14 / 87.33 & 76.10 / 59.62 & 90.40 / 79.12 \\
    \midrule
        Linear Probing  & \multirow{4}{*}{40k} & 98.94 / 91.28 & \textbf{97.23} / \textbf{90.60} & \textbf{77.60} / \textbf{73.42} & 91.26 / \textbf{85.10} \\
        Fine-tuning  & & 96.83 / 80.10 & 88.44 / 70.32 & 69.85 / 63.00 & 85.04 / 71.14 \\
        Adapter  & & 98.98 / 89.69 & 94.71 / 83.39 & 62.03 / 52.82 & 85.24 / 75.30 \\
        Prompt Tuning  & & \textbf{99.00 }/ \textbf{91.43} & 96.15 / 86.79 & 79.52 / 60.10 & \textbf{91.56} / 79.44 \\
    \midrule
        Linear Probing  & \multirow{4}{*}{60k} & 98.97 / 91.33 & \textbf{97.41} / \textbf{91.18} & \textbf{77.41} / \textbf{73.90} & \textbf{91.26} / \textbf{85.47} \\
        Fine-tuning  & & 97.08 / 79.91 & 89.05 / 69.36 & 70.59 / 62.55 & 85.58 / 70.60 \\
        Adapter  & & \textbf{99.35} / \textbf{91.74} & 95.93 / 85.07 & 64.85 / 53.75 & 86.71 / 76.85 \\
        Prompt Tuning  & & 99.29 / 91.69 & 96.47 / 85.64 & 76.94 / 57.25 & 90.90 / 78.20 \\
    \midrule
        Linear Probing  & \multirow{4}{*}{80k} & 98.94 / 91.38 & \textbf{97.31} / \textbf{90.68} & 76.69 / \textbf{73.13} & 90.98 / \textbf{85.06} \\
        Fine-tuning  & & 97.75 / 82.79 & 90.45 / 73.69 & 72.08 / 65.45 & 86.76 / 73.98 \\
        Adapter  & & \textbf{99.46} / \textbf{92.12} & 96.12 / 83.73 & 66.28 / 53.28 & 87.29 / 76.38 \\
        Prompt Tuning  & & 98.93 / 89.30 & 96.58 / 85.60 & \textbf{80.95} / 58.08 & \textbf{92.15} / 77.66 \\
    \bottomrule
    \end{tabular}%
    }
  \label{tab:trainsetsize}%
\end{table}

\subsection{Few-shot Analysis}
We now conduct experiments to investigate how participating transfer learning approaches perform when trained on extremely limited data, specifically only 640 images (320 $real$, 320 $fake$). Here, we present the results achieved by our models in a few-shot setting.

We train CLIP (ViT-Large) model using four different transfer learning strategies, i.e., (1) Linear Probing, (2) Fine-tuning, (3) Adapter Network~\cite{gao2023clip} and (4) Prompt Tuning~\cite{zhou2022learning} in a few-shot setting. We use only 32 (16 $real$ and 16 $fake$) images from each of the object categories available in the LSUN~\cite{yu2015lsun} and ProGAN~\cite{karras2017progressive} datasets. In total, we train the models using 640 $real/fake$ images. We present the achieved Average Precision (AP) and Accuracy (Acc) scores in Table~\ref{tab:32shots}. It is apparent from the results that Prompt Tuning outperforms other transfer learning strategies by a clear margin on images sampled from GAN-based, Diffusion-based and Commercial image generators. 

\subsection{Performance on Commercial Tools}
Besides evaluating the models on images sampled by a number of different GAN-based and Diffusion-based image generators, following~\cite{cozzolino2023raising} we also carry out evaluations of baseline methods, and the transfer learning strategies we employ on images generated by Commercial tools including Midjourney-V5, Adobe Firefly and DALL-E 3. We present the comparison of results in Table~\ref{tab:robustnessacrossfamilies}. The numbers clearly demonstrate that the transfer learning strategies utilized in this paper surpass previously proposed deepfake detection methods. Additionally, it's noteworthy that our models are trained using only 200k $real/fake$ images, compared to the studies we're comparing against, which utilize 720k images for training.



\begin{table}[htbp]
  \centering
  \caption{We present the results from our few-shot (32-shot) experiments, wherein we train CLIP using various transfer learning strategies on $real/fake$ images from the ProGAN dataset. We then evaluate the trained models on images generated by GANs, Diffusion models and Commercial image generators.}
  \resizebox{\linewidth}{!}{%
    \begin{tabular}{lcccc}
    \toprule
    \multirow{3}{*}{Method} & \multicolumn{3}{c}{Family of Generators} & \multirow{3}{*}{\begin{tabular}[c]{@{}c@{}}Average\\AP/Acc\end{tabular}} \\
    \cmidrule(lr){2-4} 
    & \multicolumn{1}{p{5.215em}}{\centering GAN AP/Acc} & \multicolumn{1}{p{5.215em}}{\centering Diffusion AP/Acc} & \multicolumn{1}{p{6.215em}}{\centering Comm. Tools AP/Acc} & \\
    
    \midrule
        Linear Probing  & 94.39 / 83.62 & 89.67 / 80.47 & 76.78 / \textbf{69.72} & 86.95 / 77.94 \\
        Fine-tuning  & 97.09 / 85.23 & 90.14 / 77.18 & 71.35 / 65.90 & 86.19 / 76.11 \\
        Adapter  & 97.40 / 87.27 & 90.53 / 81.12 & 61.69 / 53.93 & 83.21 / 74.11 \\
        Prompt Tuning  & \textbf{98.61} / \textbf{89.88} & \textbf{95.97} / \textbf{84.76} & \textbf{87.23} / 66.38 & \textbf{93.94} / \textbf{80.34} \\
        
    \bottomrule
    \end{tabular}%
    }
  \label{tab:32shots}%
\end{table}

\begin{table}[htbp]
  \centering
  \caption{Robustness of transfer learning strategies across different families of generative models.}
  \resizebox{\linewidth}{!}{%
    \begin{tabular}{lcccc}
    \toprule
    \multirow{3}{*}{Method} & \multicolumn{3}{c}{Family of Generators} & \multirow{3}{*}{\begin{tabular}[c]{@{}c@{}}Average\\AP/Acc\end{tabular}} \\
    \cmidrule(lr){2-4} 
    & \multicolumn{1}{p{5.215em}}{\centering GAN AP/Acc} & \multicolumn{1}{p{5.215em}}{\centering Diffusion AP/Acc} & \multicolumn{1}{p{6.215em}}{\centering Comm. Tools AP/Acc} & \\
    \midrule
        \citeauthor{wang2020cnn} (CVPR'20)  & 92.32 / 78.23 & 74.29 / 57.15 & 61.57 / 52.43 & 76.06 / 62.61 \\
        Gragn. et al. (ICME'21)  & 98.88 / 92.83 & 92.86 / 64.53 & 72.58 / 56.53 & 88.10 / 71.30 \\
        \citeauthor{corvi2023detection} (ICASSP'23)  & 99.14 / 90.67 & 86.06 / 57.15 & 66.40 / 54.62 & 83.87 / 67.48 \\
        \citeauthor{ojha2023towards} (CVPR'23)  & 95.39 / 86.13 & 93.66 / 82.77 & 75.26 / 68.42 & 88.10 / 79.11 \\

    \midrule
        Ours (LP)  & 98.22 / 90.02 & 95.60 / 86.01 & 81.95 / 72.53 & 91.92 / 82.86 \\
        Ours (FT)  & 99.32 / 89.73 & 97.72 / \textbf{92.59} & \textbf{98.52} / \textbf{94.48} & \textbf{98.52} / \textbf{92.27} \\
        Ours (Adapter)  & 99.63 / 94.13 & 96.83 / 85.70 & 70.29 / 55.17 & 88.92 / 78.33 \\
        Ours (Prompt T.)  & \textbf{99.61} / \textbf{94.16} & \textbf{97.97} / 86.86 & 85.71 / 59.62 & 94.43 / 80.21 \\
        
    \bottomrule
    \end{tabular}%
    }
  \label{tab:robustnessacrossfamilies}%
\end{table}


\section{Conclusion}
\label{sec:conclusion}

Our study examines the robustness of CLIP in detecting deepfake imagery across diverse data distributions. We explore four distinct transfer learning strategies, including Fine-tuning, Linear Probing, Prompt Tuning and training an Adapter Network, using a diverse training set of 200k images from the ProGAN dataset. Our experiments encompass evaluation on a comprehensive test set comprising 21 different image generators. 

Through our experiments, we illustrate that transfer learning strategies incorporating both the image and text components of CLIP consistently surpass the performance of simpler approaches like Linear Probing, which solely utilizes the visual aspect of CLIP. Our findings highlight Prompt Tuning's superiority over current baselines and SOTA methods, achieving significant margins of improvement while showcasing its efficacy despite minimal training parameters. Additionally, we conduct few-shot experiments, analyze robustness under post-processing operations such as JPEG compression and Gaussian blurring, and demonstrate the consistent performance of our CLIP-based detectors even with a smaller training set size of 20k images.

\begin{acks}
This research was supported by ANONYMOUS FOR REVIEW.

We extend our gratitude to the authors of~\cite{wang2020cnn, ojha2023towards, zhou2022learning, gao2023clip, cozzolino2023raising, corvi2023detection, bammey2023synthbuster} for sharing their codes, pre-trained models and collected datasets, which greatly aided our research.

We acknowledge the use of Generative AI~\cite{chatgpt} for checking and correcting the grammar of this paper. However, it is important to note that we did not use the tool to generate totally new text, rather we use it to check/correct the grammar of our provided text.
\end{acks}

\bibliographystyle{ACM-Reference-Format}
\bibliography{main}

\appendix



\end{document}